# ZNO-Eval: Benchmarking reasoning capabilities of large language models in Ukrainian

**Mykyta V. Syromiatnikov[1]**
Postgraduate student, Department of Software Engineering
ORCID: https://orcid.org/ 0000-0002-0610-3639; nik.syromyatnikov@gmail.com
**Victoria M. Ruvinskaya[1]**
PhD, Professor, Department of Software Engineering
ORCID: https://orcid.org/ /0000-0002-7243-5535; ruvinska@op.edu.ua. Scopus Author ID: 57188870062
**Anastasiya S. Troynina[1]**
PhD, Associate Professor, Department of Software Engineering
ORCID: https://orcid.org/ 0000-0001-6862-1266; anastasiyatroinina@gmail.com. Scopus Author ID: 57193992712
[1] Odesa Polytechnic National University, 1, Shevchenko Ave. Odesa, 65044, Ukraine

## ABSTRACT

As the usage of large language models for problems outside of simple text understanding or generation increases, assessing their abilities and limitations becomes crucial. While significant progress has been made in this area over the last few years, most research has focused on benchmarking English, leaving other languages underexplored. This makes evaluating the reasoning and robustness level of language models in Ukrainian particularly challenging.

The purpose of this work is to establish a comprehensive benchmark for the reasoning capabilities evaluation of large language models in the Ukrainian language. This paper presents the ZNO-Eval benchmark based on real exam tasks from Ukraine's standardized educational testing system: the External Independent Evaluation and the National Multi-subject Test. With single-answer options, multiple-choice, matching, and open-ended questions from diverse subjects, including Ukrainian language, mathematics, history, and geography, this dataset paves the way toward a thorough analysis of reasoning capabilities across different domains and complexities.

Evaluation of several well-known language models, such as GPT-3.5-Turbo, GPT-4o, GPT-4-Turbo, Mistral Large, Claude 3 Opus, and Gemini-1.5 Pro on this benchmark demonstrated the superiority of GPT-4o in both common knowledge reasoning and intricate language tasks. At the same time, Gemini Pro and GPT-4 Turbo excelled in the arithmetic domain, leading in single-answer and open-ended math problems. While all models were close to max performance in text-only common knowledge tasks like history and geography, there still is a gap for Ukrainian language and math, thus highlighting the importance of developing specialized language benchmarks for more accurate assessments of model capabilities and limitations across different languages and contexts.

This research introduced ZNO-Eval, an effective benchmark for evaluating reasoning capabilities, and thoroughly explored the abilities and limitations of modern solutions in the Ukrainian language. Future research should aim to expand the scope of ZNO-Eval to other modalities like images commonly used for exam problem description.

**Keywords:** large language model; reasoning capabilities; external independent evaluation; math; history; geography; benchmark

The recent advancements in language modeling have revolutionized the field of natural language processing, dramatically improving the context understanding in automated customer service and content generation tasks. Moreover, it was observed that large language models (LLM) could be successfully applied to solve other real-world problems outside NLP, such as making discoveries in mathematical sciences [1] or planning and executing operations with robots for a given task description in natural language [2]. Although general-purpose benchmarks for language understanding or generation with narrow homogeneous tasks are essential in baseline performance measurement and cross-model comparisons, they can barely help understand actual reasoning, complete scene-understanding capabilities, and evaluate hallucinations in complex real-world tasks where the language model makes plausible but incorrect statements.

Nowadays, while LLMs' reasoning capabilities are being actively studied, much of the current investigation and benchmarking focus disproportionately on widely spoken languages like English, leaving significant gaps in understanding how these models perform in less commonly represented languages. Ukrainian, with its rich linguistic features and growing digital presence, despite presenting a unique challenge and opportunity for such evaluation, has not received the same level of attention.

This work aims to establish a robust tool in the form of a benchmark dataset with content-rich tasks bearing real-world complexity designed to assess the reasoning capabilities and encompass the full spectrum of language understanding of publicly available large language models in the Ukrainian language.

Over the past few years, a simple increase of the trainable parameters, the so-called scaling law, ascended language models to the critical milestone of human-level performance on common general-purpose benchmarks like GLUE [3]. However, further discoveries later revealed that these metrics have a limited correlation with real-world performance and barely provide researchers with a satisfactory comprehension of actual model capabilities and limitations [4]. Since then, the benchmarks and datasets tailored to assess various understanding and reasoning tasks have significantly advanced and played a pivotal role in driving progress in language modeling.

Let us take a deeper look at widely used benchmarks focusing on reasoning capabilities:

– MMLU (Massive Multitask Language Understanding) benchmark designed to test a model's knowledge and reasoning abilities in specific subject areas includes over 57 tasks, covering multiple academic disciplines such as mathematics, history, literature, and science;

– GSM8K (Grade School Math 8K) is a standard benchmark for assessing language models' multi-step reasoning and arithmetic skills, featuring 8,000 school-level math problems;

– BIG-Bench (Beyond the Imitation Game Benchmark) is a large-scale benchmark with over 200 tasks such as logic, mathematics, common sense reasoning, and language generation, challenging models to demonstrate deeper understanding and reasoning across different cognitive tasks [5].

While the aforementioned multitask and multilingual benchmarks empower researchers with a thorough evaluation, their primary focus remains mainly on English, limiting their applicability in assessing performance across different languages, including Ukrainian. Most existing diagnostical datasets for the Ukrainian language target narrow problems like text classification or question answering. However, a good starting point for reasoning capability evaluation was recently established by the Shared Task on Fine-Tuning Large Language Models for Ukrainian. This task contains almost 4,000 machine-readable questions and answers from the Ukrainian External Independent Evaluation (EIE) exam, covering two subjects: the history of Ukraine and the Ukrainian language and literature [6]. Compared to other benchmarks, this dataset better reflects real-world complexity since its origin serves as a critical measure of academic proficiency for school graduates across the nation. However, this benchmark has three key disadvantages: it only contains questions with one correct answer; it does not evaluate arithmetic skills; and tasks are not grouped by tests, thus making comparison with human performance harder.

This paper introduces ZNO-Eval, a comprehensive benchmark designed to assess the reasoning capabilities of large language models in the Ukrainian language. ZNO-Eval is inspired by the structure and content of the standardized Ukrainian educational testing system, the External Independent Evaluation. By leveraging the question format defined by the ZNO dataset, a dataset with diverse subjects such as language, mathematics, history, and geography was created.

Fig. 1 illustrates the sample task schema designed and used for all tasks and exams.

In addition, we present the results of evaluating several well-known state-of-the-art large language models on the established benchmark. Evaluations were performed in a zero-shot prompting manner via the UA-LLM framework [7]. The prompt instructs the model to output the correct answer letter/number, a sequence of letters/numbers, or a calculated result. Only the first entry is used if the output contains more answers than needed. All visual tasks with images as part of a question or answer options were either replaced with text descriptions or skipped if such translation was impossible.

```
1 {
2     "task_id": 15,
3     "question": "До кожного початку речення (1–3) доберіть його закінчення (А – Д) так, щоб
                   утворилося правильне твердження.",
4     "answers": [
5         {"answer": "1", "text": "Функція y=√(x−4)"},
6         {"answer": "2", "text": "Функція y=2"},
7         {"answer": "3", "text": "Функція y=x^3"},
8         {"answer": "А", "text": "спадає на проміжку (−∞;0)."},
9         {"answer": "Б", "text": "не визначена в точці x=1."},
10        {"answer": "В", "text": "набуває від'ємного значення в точці x=8."},
11        {"answer": "Г", "text": "набуває додатного значення в точці x=−3."},
12        {"answer": "Д", "text": "є непарною."}
13    ],
14    "answer_vheader": ["А", "Б", "В", "Г", "Д"],
15    "answer_hheader": ["1", "2", "3"],
16    "correct_answer": ["Б", "Г", "Д"],
17    "comment": "Коментар\nТЕМА: Алгебра і початки аналізу. Функції.",
18    "with_photo": false
19 }
```

*Fig. 1.* **ZNO-Eval data format: sample math task with multiple answers**

The Ukrainian language and literature dataset consists of 49 exams administered over the past ten years, containing 2.746 questions in total. To evaluate Ukrainian language proficiency, we selected four National Multi-subject Test (NMT) exams, each containing 30 tasks, resulting in 120 questions. These exams were chosen intentionally, as in contrast to EIE, postponed after 2021, its temporary replacement – NMT, does not contain open-ended tasks requiring subject matter expert assessment. Fig. 2 presents a graph showing the average test scores achieved by each model.

Each graph's "max" bar represents the distribution of exam points per task type. In contrast, the "max solvable" bar can be interpreted as a practical ceiling for these evaluations, as it only considers tasks possible to solve with text-only input. For each single correct answer task, four or five answer options are presented, of which only one is correct. Each matching task consists of information indicated by numbers and letters. To complete the task correctly, matching the information marked with numbers and letters (to form logical pairs) is necessary.

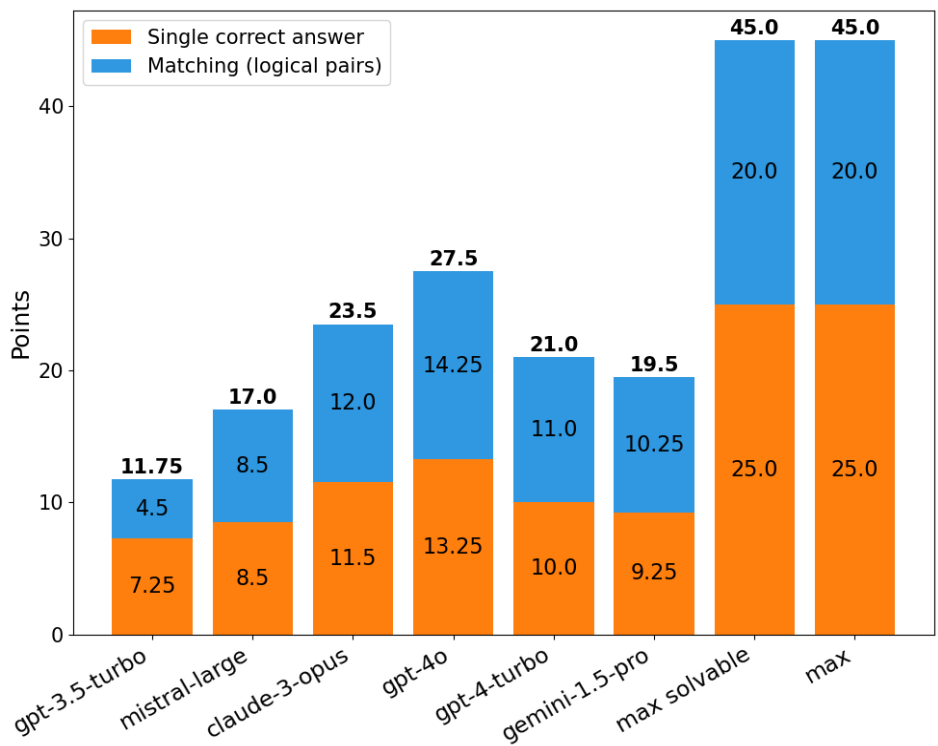

*Fig. 2.* **Average results of Ukrainian language assessments**

A dataset of 3 EIE and 6 NMT math exams was collected for arithmetic reasoning evaluation. Combining single correct answer tasks, matching, and open-ended questions requiring multi-step reasoning results in 230 entries. Due to the same reason as for the Ukrainian language evaluation, the four NMT exams with 30 questions each were selected. Additionally, all formulas in question and answer options have been converted to either plain text or LaTeX formatting, depending on the formula complexity. This adjustment was necessary as the original MathML format significantly increased the total number of input tokens. Math evaluation results are demonstrated in Fig. 3.

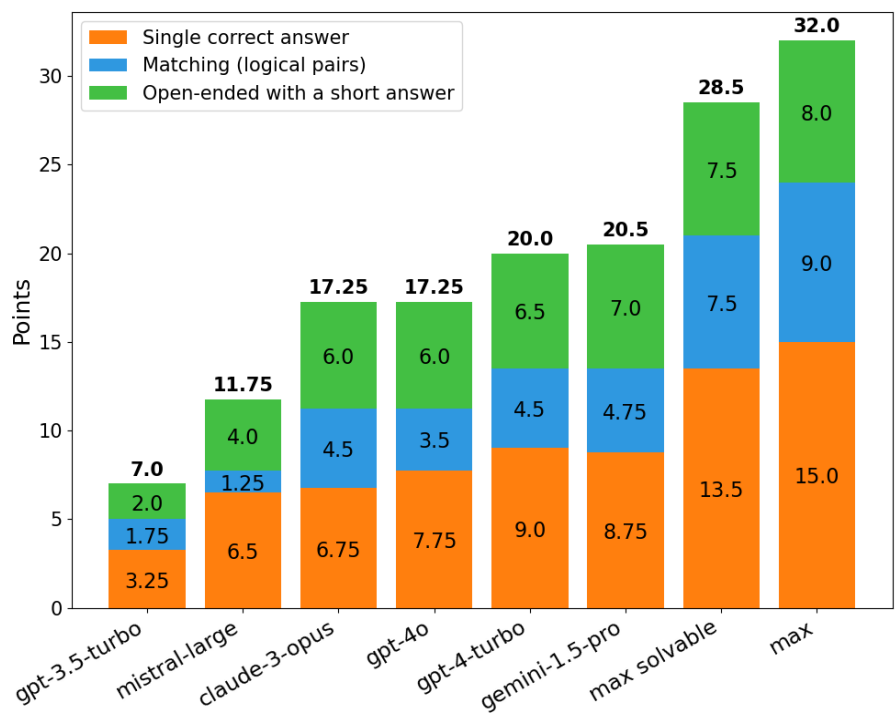

*Fig. 3.* **Average results of math assessments**

The History of Ukraine dataset was compiled from NMT and EIE exams, resulting in 48 tests and 2.640 questions, covering key historical periods, significant events, and influential figures that have shaped the country. Fig. 4 demonstrates the evaluation result, for which a subset of the dataset with four NMT tests and 120 questions was selected.

In addition to single correct answer and matching questions used in math and language exams, the history tests introduce two more types. For sequencing tasks, a list of events must be arranged chronologically, while tasks with three correct answers offer seven answer options.

The Geography dataset was carefully assembled from 32 External Independent Evaluation tests, totaling 1,788 questions. This dataset covers a broad spectrum of geographical topics, providing a well-rounded assessment of common knowledge. Fig. 5 presents the evaluation results for a subset of this dataset, encompassing three exams with 54 tasks each and 162 questions in total.

The evaluation of large language models using the proposed ZNO-Eval benchmark reveals that GPT-4o performed the best overall, demonstrating strong reasoning and comprehension capabilities across various subjects. However, Gemini-1.5 Pro and GPT-4 Turbo outperformed the leader in handling complex, unstructured arithmetic reasoning tasks, showing distinct strengths in specific areas. The models performed well on the NMT subsets for history and geography, highlighting their proficiency in factual recall and common knowledge. They also demonstrated strong capabilities in handling structured tasks and schemas in a zero-shot manner. However, the limitations in understanding more nuanced language and terms led to struggles with single-answer and matching questions for Ukrainian language and math exams, while these tasks are ordinary for most examinees.

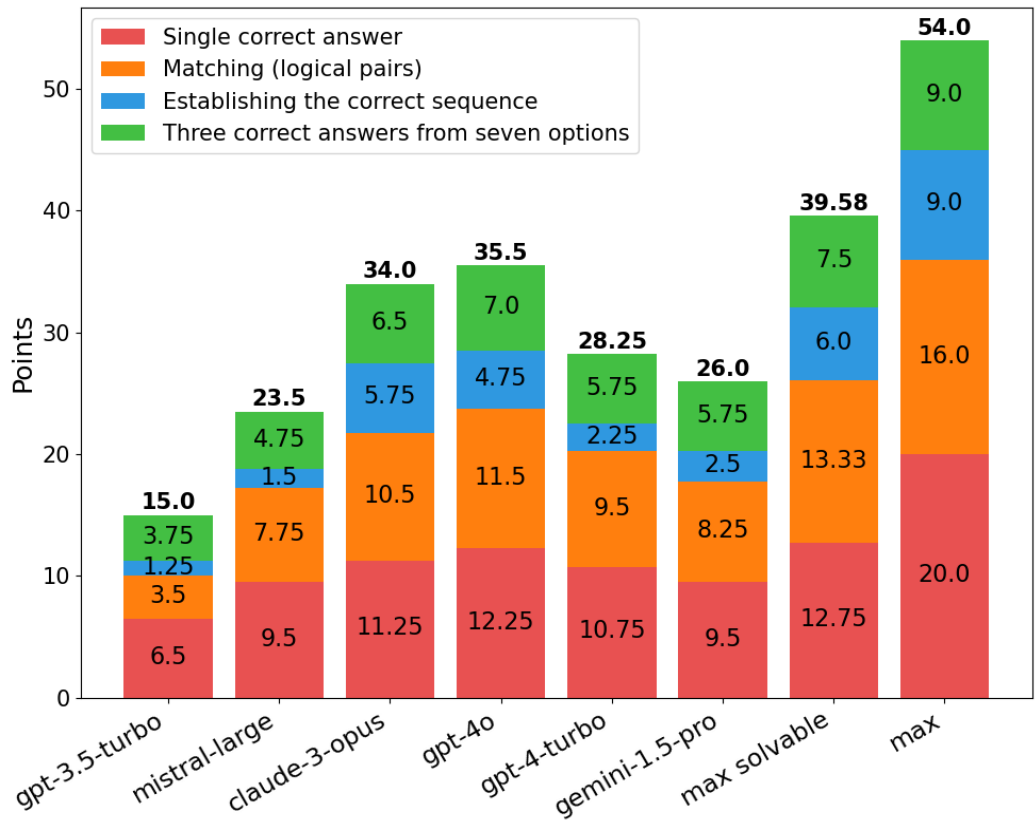

*Fig. 4.* **Average results of history assessments**

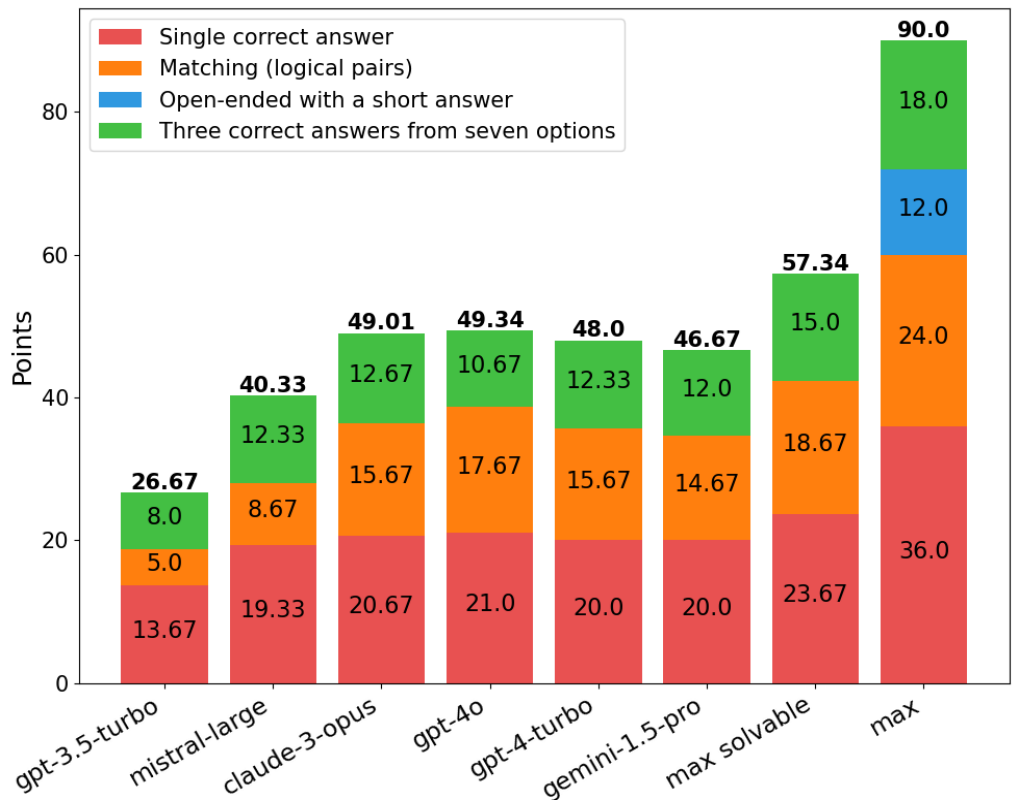

*Fig. 5.* **Average results of geography assessments**

These findings underscore the importance of creating language-specific benchmarks to ensure that LLMs can perform effectively across diverse languages and domains. The results from the ZNO-Eval benchmark provide valuable insights into the current state of LLMs' reasoning abilities and limitations in the Ukrainian language, while the established benchmark [8] itself contributes to the broader goal of building general-purpose AI systems that can serve diverse communities.

# ZNO-Eval: Оцінка розумових здібностей великих мовних моделей при роботі з україномовними текстами

**Сиром'ятніков Микита Валерійович**[1)]
Аспірант каф. Інженерії програмного забезпечення
ORCID: https://orcid.org/0000-0002-0610-3639; nik.syromyatnikov@gmail.com
**Рувінська Вікторія Михайлівна**[1)]
Канд. техніч. наук, професор каф. Інженерії програмного забезпечення
ORCID: https://orcid.org/0000-0002-7243-5535; ruvinska@op.edu.ua. Scopus Author ID: 57188870062
**Тройніна Анастасія Сергіївна**[1)]
Канд. техніч. наук, доцент каф. Інженерії програмного забезпечення
ORCID: https://orcid.org/0000-0001-6862-1266; anastasiyatroinina@gmail.com. Scopus Author ID: 57193992712
[1)] Національний університет «Одеська політехніка», пр. Шевченка, 1. Одеса, 65044, Україна

## АНОТАЦІЯ

Оскільки усе частіше великі мовні моделі використовуються для вирішення завдань, що виходять за рамки простого розуміння та генерації тексту, оцінка їхніх можливостей та обмежень стає критично важливою. Хоча в цьому напрямку було досягнуто значного прогресу за останні кілька років, більшість досліджень зосереджено на тестуванні англійської мови, залишаючи інші мови недостатньо дослідженими. Це робить оцінку розумових здібностей та стійкості мовних моделей для української мови особливо складною задачею.

Метою цієї роботи є створення дігностичного набору для оцінки розумових здібностей великих мовних моделей у українській мові. У цій роботі представлено датасет ZNO-Eval, що базується на завданнях з української системи стандартизованого освітнього тестування: зовнішнього незалежного оцінювання та національного мультипредметного тесту. Утворений набір, що включає запитання з однією або декількома відповідями, задачі на відповідність, а також відкриті питання з української мови, математики, історії та географії, прокладає шлях до всебічного аналізу розумових здібностей мовних моделей у різних галузях та з різними рівнями складності.

Оцінка відомих мовних моделей, таких як GPT-3.5-Turbo, GPT-4o, GPT-4-Turbo, Mistral Large, Claude 3 Opus та Gemini-1.5 Pro на побудованому діагностичному наборі продемонструвала перевагу GPT-4o у завданнях, що потребують загальних знань, а також у складних мовних задачах. У той же час, Gemini Pro і GPT-4 Turbo досягли найкращих результатів у арифметичних завданнях, випередивши конкурентів у математичних запитаннях з одним правильним варіантом та відкритою відповіддю. Хоча всі моделі досягли практично максимально можливих результатів у тестуванні загальних знань, що включає історію та географію, існує значний розрив для тестів з української мови та математики – це підкреслює важливість розробки спеціалізованих датасетів для більш точної оцінки можливостей та обмежень моделей у різних мовах і контекстах.

У рамках цієї роботи було представлено ZNO-Eval - ефективний датасет для оцінки розумових здібностей, а також було детально досліджено можливості та обмеження сучасних рішень для української мови. Майбутні дослідження включатимуть розширення ZNO-Eval на інші модальності, такі як зображення, що використовуються для опису тестових запитань.

**Ключові слова:** великі мовні моделі; розумові здібності; зовнішнє незалежне оцінювання, математика; історія; географія; діагностичний набір